# Industrial Internet Robot Collaboration System and Edge Computing Optimization

Haopeng Zhao[1a*†], Dajun Tao[2b†], Tian Qi[3c], Jingyuan Xu[4d], Zijie Zhou[5e], and Lipeng Liu[6f]
[1] Independent Researcher, Mountain View, CA, USA
[2] School of Engineering, Carnegie Mellon University, Pittsburgh, PA, USA
[3] College of Arts and Sciences, University of San Francisco, San Francisco, CA, USA
[4] School of Computer and Information Science, University of the Cumberlands, Williamsburg, KY, USA
[5] College of Engineering, Northeastern University, Boston, MA, USA
[6] College of Engineering, Peking University, Beijing, China
[a*] haopeng.zhao1894@gmail.com, [b] dajunt@alumni.cmu.edu, [c] tqi51212@gmail.com, [d] jxu65428@ucumberlands.edu, [e] zhou.zijie@northeastern.edu, [f] lipeng.liu@pku.edu.cn
†These authors contributed equally to this work and are co-first authors.

***Abstract*—In industrial-Internet environments, mobile robots must generate collision-free global routes under stochastic obstacle layouts and random perturbations in commanded linear and angular velocities. This paper models a differential-drive robot with nonholonomic constraints, then decomposes motion into obstacle avoidance, target turning, and target approaching behaviors to parameterize the control variables. Global path planning is formulated as a constrained optimization problem and converted into a weighted energy function that balances path length and collision penalties. A three-layer neural network represents the planning model, while simulated annealing searches for near-global minima and mitigates local traps. During execution, a fuzzy controller uses heading and lateral-offset errors to output wheel-speed differentials for rapid correction; edge-side computation is discussed to reduce robot–server traffic and latency. Matlab 2024 simulations report deviation within ±5 cm, convergence within 10 ms, and shorter paths than two baseline methods. The approach improves robustness of global navigation in practice.**

## I. INTRODUCTION

Industrial Internet scenarios such as warehousing, inspection, and flexible manufacturing increasingly rely on mobile robots. With advances in artificial intelligence and high-speed communications [1], robots can integrate sensing, planning, and control more tightly, yet cluttered environments and disturbances in commanded linear and angular velocities still cause tracking deviations and collision risk.

Path planning is therefore a core topic in mobile robotics. Many approaches optimize objectives such as minimum distance or minimum energy consumption [2], but high-precision navigation in complex environments remains difficult when local minima and execution disturbances are present. A practical solution must respect nonholonomic motion constraints while maintaining collision avoidance and stable tracking.

Edge computing can improve responsiveness by moving computation closer to the robot, reducing robot-server traffic and latency and enabling timely sharing of environmental information among collaborative robots. This paper studies global path planning and global path control by combining an energy-based planner with an error-correcting controller. Specifically, we establish a robot model and decompose motion into obstacle avoidance, target turning, and target approaching behaviors; simulated annealing is used to search low-energy paths, and a fuzzy controller outputs wheel-speed differentials from heading and lateral-offset errors. Matlab 2024 simulations compare the proposed method with two baselines..

## II. DYNAMIC ANALYSIS OF MOBILE ROBOT

### *A. Dynamic Model*

Set the wheel radius of the mobile robot as r and the distance between the drive shafts as l. The pose matrix is defined in (1), where (x, y) gives the robot position and θ gives the heading direction.

$$p = \left[x, y, \theta\right]^{T} \tag{1}$$

Based on the above pose, the dynamic equation of the mobile robot is established in (2) [3]. In (2), $\mu_1$ and $\mu_2$ are control inputs, $c_1$ and $c_2$ are inertia terms, $\Delta g_x$, $\Delta g_\gamma$ and $\Delta g\theta$ are nonlinear functions, γ is the Lagrange multiplier, and n is the robot mass. The last line in (2) denotes the non-holonomic constraint. According to the above calculation results, the dynamic model is obtained and shown in Fig. 1.

$$\begin{cases} \ddot{x} = \dfrac{\gamma}{n}\sin\theta + c_1\mu_1\cos\theta + \Delta g_x\left(p, \dot{p}\right) \\ \ddot{y} = \dfrac{\gamma}{n}\sin\theta + c_1\mu_1\cos\theta + \Delta g_\gamma\left(p, \dot{p}\right) \\ \ddot{\theta} = c_2\mu_2 + \Delta g_\theta\left(p, \dot{p}\right) \\ \dot{x}\sin\theta - \dot{y}\sin\theta = 0 \end{cases} \tag{2}$$

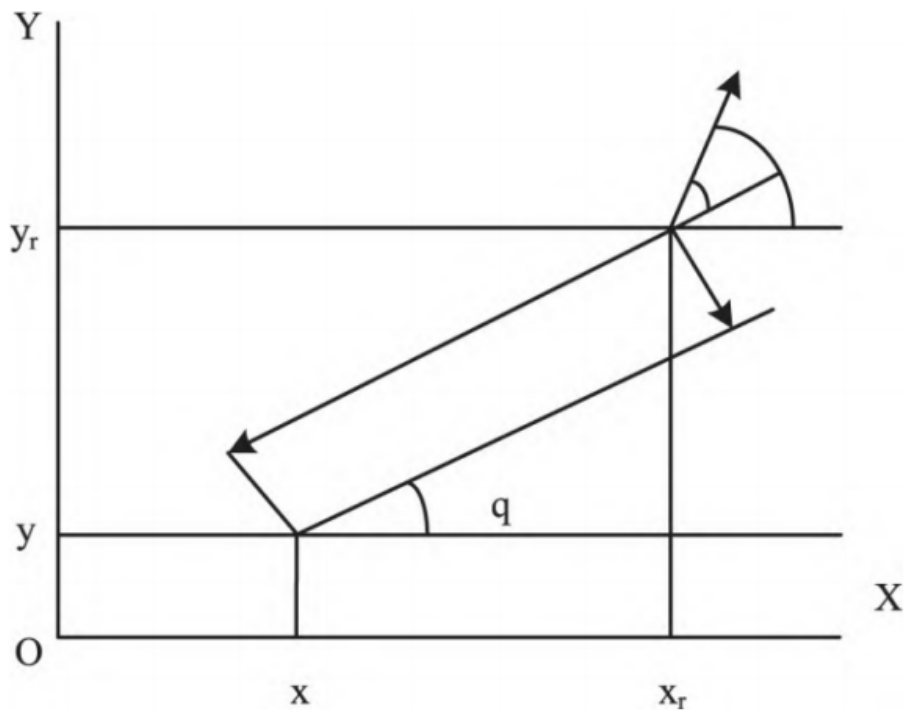

Fig. 1. Dynamics model of the mobile robot

Since the number of balance wheels does not affect the robot dynamics, the model is updated to the kinematic form in (3), where η denotes linear velocity and υ denotes angular velocity.

$$\begin{cases} \dot{x} = \eta\cos\theta \\ \dot{y} = \eta\sin\theta \\ \dot{\theta} = v \end{cases} \tag{3}$$

## B. *Analysis of Robot Motion Behaviors*

To improve controllability, robot motion is decomposed into three behaviors parameterized by linear velocity v and angular velocity ξ. Obstacle avoidance drives ξ using the deviation angle β and obstacle bearing φ, while v is adjusted by the minimum obstacle distance d_obstacle, as in (4).

$$\begin{cases} \xi = -\beta \bullet \rho_{\xi}\left(|\ \varphi| - \pi / 2\right) \\ v = \rho_{v} / d_{obstacle}^{m} + \varepsilon_{\xi} \end{cases} \tag{4}$$

Target approaching (positioning) limits v by the goal distance d_goal and the maximum speed v_max, while still considering obstacle-related terms, as in (5).

$$v = \min\left(o_{goal} d_{goal}, v_{max}, \rho_{v} / d_{obstacle}^{m} + \varepsilon_{v}\right) \tag{5}$$

Target turning sets ξ from the difference between φ_goal and φ_obstacle, and bounds v by v_tmax, as in (6).

$$\begin{cases} \xi = \beta \bullet \rho_{\xi}\left(|\ \phi_{goal}| - \varphi_{goal}\right) \\ v = \min\left(o_{goal} d_{goal}, v_{tmax}, \rho_{v} / d_{obstacle}^{m} + \varepsilon_{v}\right) \end{cases} \tag{6}$$

# III. GLOBAL PATH CONTROL METHOD OF ROBOT BASED ON FUZZY CONTROL

Based on the analysis results of the robot's motion behaviors, global path planning is carried out for the robot. The fuzzy control method [5] is introduced to promptly correct the path errors of the robot, thereby achieving the planning and control of the global path of the mobile robot.

## A. *Global Path Planning*

According to the analysis results of the robot's motion behaviors obtained in Section 2, artificial neural network principles are used to establish the global path planning model, and simulated annealing is applied to search an optimal path in a complex environment. The planning model is illustrated in Fig. 2.

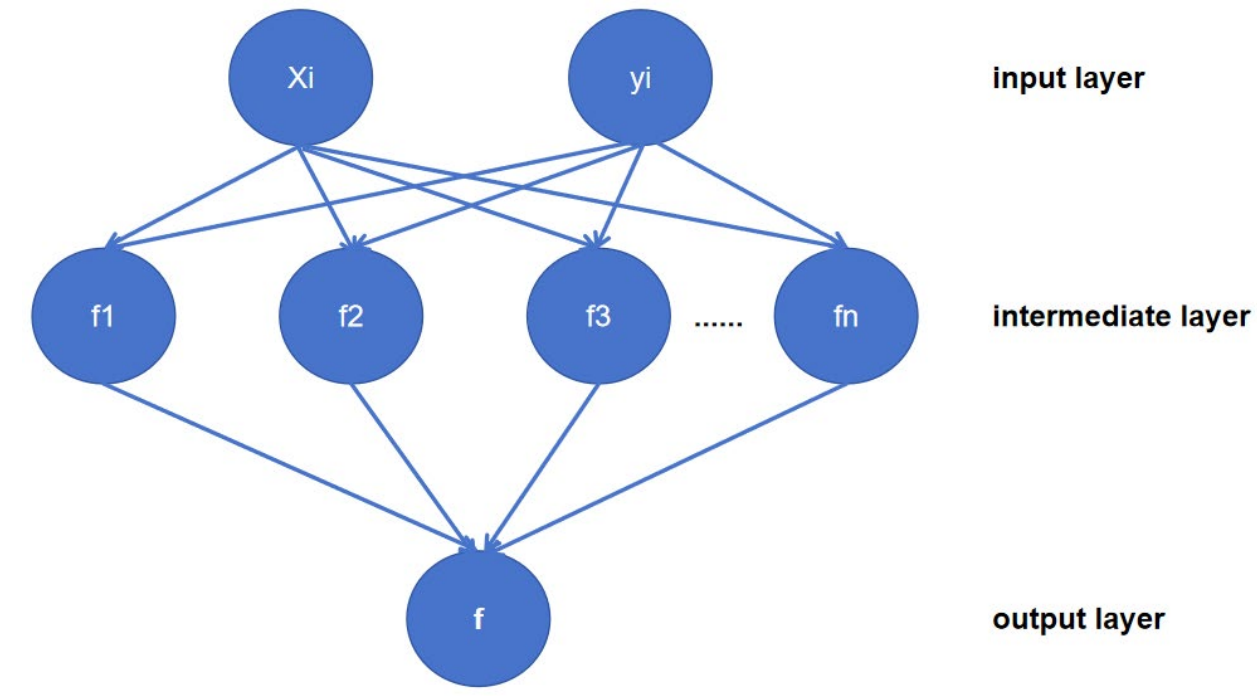

Fig. 2. Global path planning model of the mobile robot

The global path planning model is formulated as a constrained optimization problem, as shown in (7), where g(x) is the objective function and h_i(x) are inequality constraints.

$$\begin{cases} \min g\left(x\right), x \in R^{n} \\ s \bullet t.h_{i}\left(x\right) \leq 0 \end{cases} \tag{7}$$

To improve solvability, the model is converted into an energy function F that includes a path-length term F_l and a collision-penalty term F_z, as defined in (8).

$$\begin{cases} F_{l} = \sum_{i=1}^{M-1} L_{i}^{2} = \sum_{i=1}^{M-1} [(x_{i+1} - x_{i})^{2} - (y_{i+1} - y_{i})^{2}] \\ F_{z} = \sum_{i=1}^{M} \sum_{k=1}^{K} Z_{i}^{k} \end{cases} \tag{8}$$

A weighting coefficient δ_l is introduced to obtain the final energy function and its minimum value, as shown in (9), where δ_l balances F_l and F_z. Here, M is the number of path points, K is the number of obstacles, and Z_i^k represents the penalty between obstacle k and path point q(x_i, y_i).

$$\begin{cases} F = \delta_{l} F_{l} + \left(1 - \delta_{l}\right) F_{z} \\ \min F = \delta_{l} F_{l} + \left(1 - \delta_{l}\right) F_{z} \end{cases} \tag{9}$$

To solve the established planning model, simulated annealing iteratively generates neighboring paths, evaluates the energy change ΔF, and accepts inferior solutions with a certain probability to avoid local minima. As the temperature decreases, the search converges to a low-energy global path.

## B. *Global Path Control Method of the Robot*

According to the detection range of the sensors on the mobile robot and the moving environment, the angle deviation is set to ±10° and the central deviation does not exceed ±100 mm. When the deviation exceeds the threshold, the robot is judged to have deviated from the track, and the fuzzy controller is used to regulate the wheel-speed difference to keep the robot moving safely on the selected path.

When designing the controller, the input and output fuzzy sets are defined first (inputs: angle deviation and central deviation; output: left–right wheel-speed difference). A triangular membership function is used for error correction, as shown in (10).

$$\rho_{A}(x)=\begin{cases}\dfrac{x-d}{c-d}, c\geq x\geq d \\ \dfrac{e-x}{e-c}, e\geq x>c \\ 0, d>x/e>x\end{cases} \tag{10}$$

Based on expert rules, fuzzy inference is performed, and the fuzzy output is defuzzified to obtain a crisp control action, as shown in (11).

$$J_0=\frac{\sum_{i=1}^{m}\left(\omega_c\left(J_i\right)\times J_i\right)}{\sum_{i=1}^{m}\omega_c\left(J_i\right)} \tag{11}$$

## IV. EXPERIMENT

Matlab 2024 was used to simulate global path control; the simulated robot and environment are shown in Fig. 3 and Fig. 4. Three methods were compared: the proposed deep-learning-based global path control, a visual-servo generalized-constraint planner, and a dynamic/static safety-field planner.

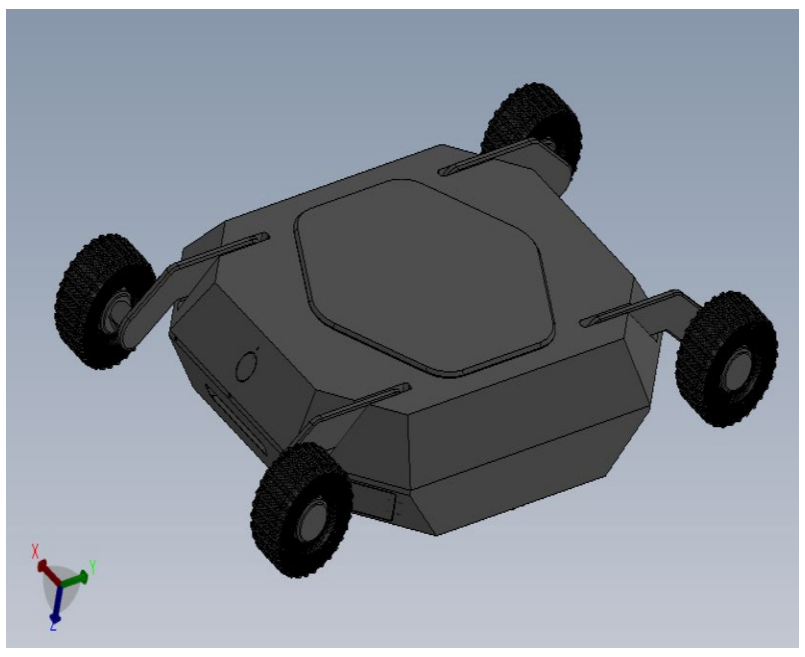

Fig.3. A simulated mobile robot

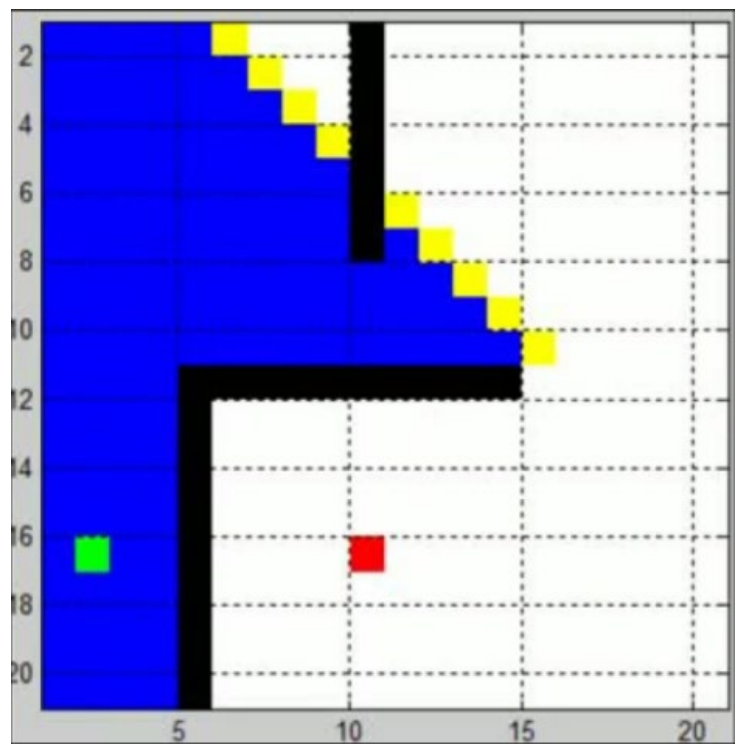

Fig. 4. Simulation path map

### A. *Comparison of the angular deviations of the robot paths*

With increasing control time, all methods exhibited deviation, but the proposed method produced the smallest path deviation and stabilized the deviation convergence fastest, attributed to the prior motion-behavior analysis.

### B. *Comparison of the lengths of the paths planned by the robot*

As iterations increased (100–500), path lengths decreased for all methods (Table 1); the proposed method achieved the shortest mean path (114 m) and the best result at 500 iterations (106 m), compared with 132 m (visual-servo) and 130 m (safety-field) mean lengths.

Table 1. path planning length test results for different methods

| **Number of iterations / times** | **The proposed method** | **Ppath planning method for robots based on visual servo generalized constraint** | **Robot path planning method based on dynamic and static safety field** |
|---|---|---|---|
| 100 | 124 | 143 | 144 |
| 200 | 118 | 139 | 138 |
| 300 | 113 | 135 | 130 |
| 400 | 109 | 125 | 121 |
| 500 | 106 | 118 | 117 |

## V. CODE EXAMPLE OF GLOBAL PATH PLANNING USING NEURAL NETWORK ALGORITHM

The following is to present each part of the code separately, and add detailed explanations after each piece of code:

### A. *Import necessary libraries*

Import numpy, TensorFlow, and skfuzzy.control to support arrays, neural planning, and fuzzy correction.

### B. *Modeling of the dynamic equation of the mobile robot*

MobileRobotDynamics stores r, l, pose=[x,y,theta]; update_pose(v,w,dt) updates pose using trigonometric relations.

### C. *Global path planning model based on neural network*

build_path_planning_model builds Sequential Dense(64,relu), Dense(64,relu), Dense(2,linear) and compiles with Adam/MSE.

### D. *Design of the fuzzy controller*

Define angle_deviation∈[-10,10], center_deviation∈[-100,100], output speed_difference; use triangular membership and fuzzy rules.

### E. *Main function, simulating the operation process of the robot*

Initialize dynamics, model, and controller; set dt=0.1, num_steps=100; loop predict, update pose, defuzzify, print.

## VI. CONCLUSION

This paper studied global path planning and global path control for a mobile robot in complex environments. A behavior-based dynamic analysis was used to parameterize obstacle avoidance, target turning, and target approaching; an energy-function planner combined neural-network modeling with simulated annealing to reduce collision risk and avoid local minima; and a fuzzy controller corrected heading and center deviations (±10° and ±100 mm) by regulating differential wheel speed.

Matlab 2024 simulations (Fig. 3–Fig. 4) show that the proposed method produces smaller path deviations (within ±5 cm) and faster deviation convergence (about 10 ms) than two baseline planners, while also generating shorter planned paths. Table 1 reports the best path length at 500 iterations (106 m) and the smallest mean length (114 m), compared with longer paths from the visual-servo and safety-field methods.

Limitations include dependence on modeling assumptions, hand-crafted fuzzy membership functions/rules, and sensitivity to sensor noise or abrupt environmental changes. Future work will incorporate uncertainty-aware sensing, adaptive tuning of fuzzy parameters, and real-robot experiments to validate robustness and real-time performance.